\documentclass[10pt,twocolumn,letterpaper]{article}

\usepackage{cvpr}
\usepackage{times}
\usepackage{epsfig}
\usepackage{graphicx}
\usepackage{amsmath}
\usepackage{amssymb}

\newtheorem{prop}{Proposition}

\usepackage[pagebackref=true,breaklinks=true,letterpaper=true,colorlinks,bookmarks=false]{hyperref}

\cvprfinalcopy 

\def\cvprPaperID{4229} 

\ifcvprfinal\fi
\begin{document}

\title{Mean Local Group Average Precision (mLGAP): A New Performance Metric for Hashing-based Retrieval}

\author{Pak Lun Kevin Ding\thanks{Indicates equal contributions.},
Yikang Li\footnotemark[1],
Baoxin Li\\
School of Computing, Informatics, and Decision Systems Engineering\\
Arizona State University\\
{\tt\small \{kevinding, yikangli, baoxin.li\}@asu.edu}
}

\maketitle

\begin{abstract}
The research on hashing techniques for visual data is gaining increased attention in recent years due to the need for compact representations supporting efficient search/retrieval in large-scale databases such as online images. Among many possibilities, Mean Average Precision(mAP) has emerged as the dominant performance metric for hashing-based retrieval. One glaring shortcoming of mAP is its inability in balancing retrieval accuracy and utilization of hash codes: pushing a system to attain higher mAP will inevitably lead to poorer utilization of the hash codes. Poor utilization of the hash codes hinders good retrieval because of increased collision of samples in the hash space. This means that a model giving a higher mAP values does not necessarily do a better job in retrieval. In this paper, we introduce a new metric named Mean Local Group Average Precision (mLGAP) for better evaluation of the performance of hashing-based retrieval. The new metric provides a retrieval performance measure that also reconciles the utilization of hash codes, leading to a more practically meaningful performance metric than conventional ones like mAP. To this end, we start by mathematical analysis of the deficiencies of mAP for hashing-based retrieval. We then propose mLGAP and show why it is more appropriate for hashing-based retrieval. Experiments on image retrieval are used to demonstrate the effectiveness of the proposed metric.
\end{abstract}

\section{Introduction}

\begin{figure*}
	\begin{center}
		\includegraphics[width=0.8\linewidth]{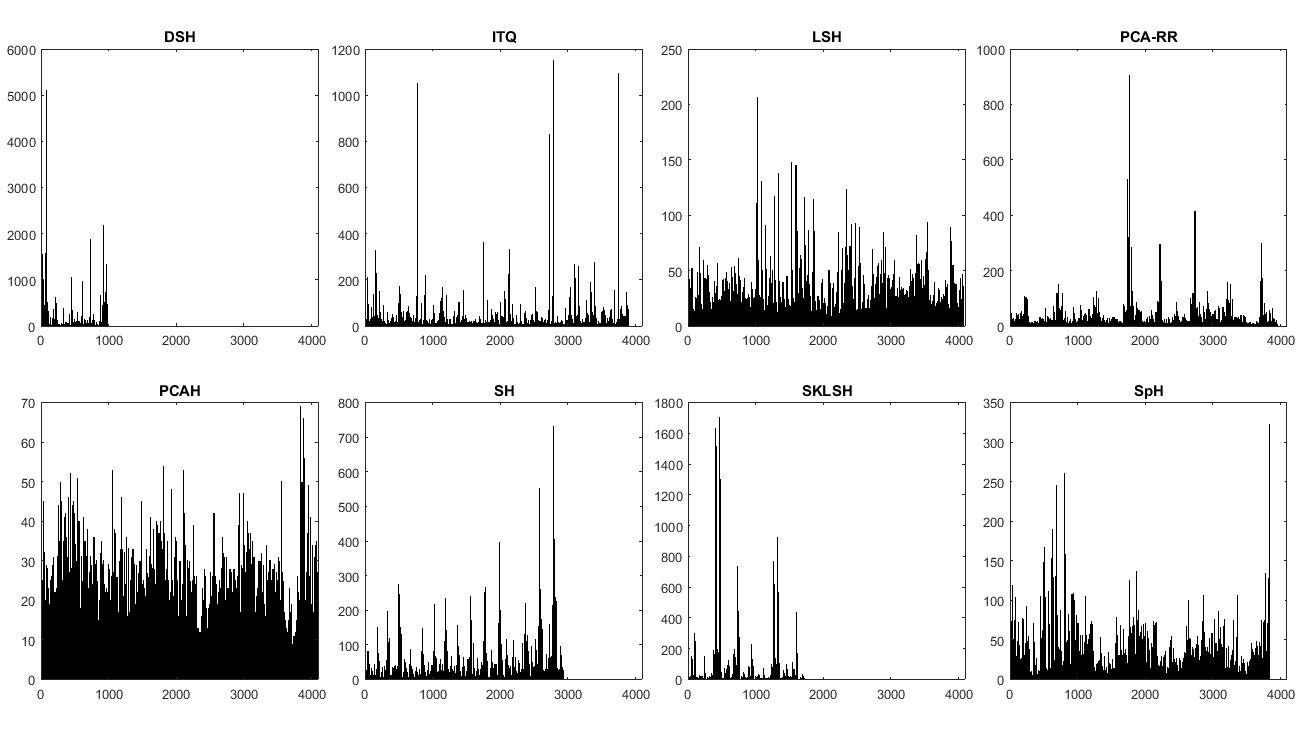}
	\end{center}
	\caption{The figures show the histograms of the 12-bit hash codes from different hash models on CIFAR10 \cite{cifar10}. (Top) From left to right: DSH\cite{densitysh}, ITQ\cite{itq}, LSH\cite{lsh}, PCA-RR\cite{itq}; (Bottom) from left to right: PCAH\cite{pcah}, SH\cite{sh}, SKLSH\cite{sklsh}, SpH\cite{sph}.
	}
	\label{fig:usage}
\end{figure*}

In recent years, due to the rapid growth of images and videos on the Internet, searching for relevant and similar images or videos from the Web has become a very practical and challenging task. 
Many retrieval methods, including text-based search or content-based retrieval, have been proposed. Content-based retrieval is desired in many scenarios since it does not require the availability of metadata ($e.g.$, tags or annotations).
However, content-based approaches are in general computationally costly, due to both the excessive computation needed for obtaining content descriptors and the high cost associated with searching for matches from a large-scale dataset using the descriptors. 

To address both the time (speed) and the space (memory requirement) complexities facing large-scale applications, researchers have proposed various approximate techniques, and in particular the hashing approaches \cite{hash2,hash1} that map input images to very compact binary codes so that comparison/matching of any pair of images can be approximately done by using their corresponding binary hash codes, without invoking the original image or descriptors. Since all images are represented in compact binary vectors (usually shorter than a few hundreds of bits per image), not only the amount of storage/memory, but also the computational time for a comparison operation can be greatly reduced.

There are several performance metrics that are commonly used to evaluate the performance of a hashing-based retrieval system. These include Precision at $k$ samples ($P@k$), Precision at Hamming distance ($P@d_H\leq r$), and Mean Average Precision (mAP). We argue that such metrics do not provide a comprehensive measure for hashing-based retrieval performance. For example, both mAP and Precision can only indict the accuracy in retrieving relevant samples but fail to provide information on how well the hash codes are distributed. The distribution of the hash codes from a hashing scheme is of importance to the retrieval performance, since, for example, high collision (or low utilization of the hash space) is detrimental to good retrieval. In particular, pushing a system to attain higher mAP, which is currently the dominant metric used in the literature, will inevitably lead to poorer utilization of the hash space. This explains why some unsupervised approaches ($e.g.$,\cite{pcah}) usually have a high utilization of the binary codes after hashing, but do not give good performance numbers in mAP for large-scale datasets ($e.g.$, usually the mAP values are below $20\%$ with 12-bit hash codes for the CIFAR-10 dataset). Some supervised hashing methods take advantages of label information \cite{cnnh, dsh, dnnh} to gain better retrieval results in mAP. However, with the label information used in the training stage, more samples from the same category are encoded into the {\sl same} binary code, $\ie$, high collision in the hash space. Even though the final performance in terms of mAP is much better ($e.g.$, in \cite{dsh}, the mAP value with 12-bit hash codes for the CIFAR-10 dataset is more than $60\%$) compared with unsupervised hashing approaches, it is questionable to accept this as true improvement, because of the high collision, since one may argue that, in the extreme case, hashing becomes a classification problem with only ($e.g.$) 10 unique hashing codes for ($e.g.$) 10 classes, and then it is no longer possible to rank-order the samples in the retrieval process. 

Consequently, in general it is reasonable to argue that, for hashing-based retrieval, seeking high values in mAP would lead to failure in forming hash codes as much dispersed as possible, and thus could push the retrieval task towards a classification-like task. This would essentially beat the purpose of doing hashing for retrieval. Therefore, it is desired to have some new performance metric that takes into consideration both retrieval accuracy and dispersion of the hashing codes.

In this work, we propose a new metric for evaluating hashing-based retrieval. This new metric is derived from conventional mAP and Precision, while keeping in mind the utilization of the hash space, and hence it has the potential of gaining the benefits of both ends. Furthermore, we will show that the new metric would not be affected by the order of the retrieved samples when the collision of hash codes happens, whereas mAP suffers from significant differences if the order of retrieved samples changes.

The main contribution of this work is three-fold:
\begin{itemize}
    \item We propose a new metric that more meaningfully evaluates hashing-based retrieval for its balanced consideration of accuracy and utilization of the hash space.
    \item We provide mathematical foundations for the new metric through analyzing deficiencies of mAP in terms of its impact on utilization of the hash space.
    \item We demonstrate that when the samples are mapped more uniformly to the binary space, there are rooms to preserve the similarity of the samples in the visual or semantic domain.
\end{itemize}

The rest of the paper is organized as follows:
In section \ref{sec:related} we discuss the related work.
After that, we introduce some popular metrics for measuring the performance of retrieval system in Section \ref{sec:metric}.
We then describe our proposed metric in Section \ref{sec:proposed}.
In Section \ref{sec:exp}, we demonstrate some experimental results, and end with conclusions in Section \ref{sec:conc}.

\section{Related Work} \label{sec:related}
Many hashing methods have been proposed for data-intensive applications in machine learning,
computer vision, data mining and related areas~\cite{hash2, hash1}. 
One key objective of hashing in such applications is to encode an input vector, usually high-dimensional,
to a compact binary vector, while preserving some similarity measure of the original data.
As it is in general much more efficient to compare a pair of binary codes ($e.g.$ using the Hamming distance) than doing the same with the original vector data, hashing has become a useful technique for applications involving large-scale datasets like image retrieval on the Internet.

Hashing techniques can be divided into two categories: {\em data-independent} methods, and {\em data-dependent} methods. One representative {\em data-independent} approach is Locality-Sensitive Hashing (LSH)~\cite{lsh},
which uses random projections to generate hash functions. 
LSH has been extended to several versions, such as Kernelized LSH~\cite{lsh1} and other variants~\cite{lsh3, lsh2}.
However, empirical results suggest that, LSH and other data-dependent approaches like \cite{dd} usually require long bit length to maintain high precision and recall. This not only lowers the speed performance,
but also increases space complexity. Hence data-dependent approaches are not deemed as the best option for large-scale problems like Internet-scale image retrieval.

On the other hand,
{\em data-dependent} approaches are supposed to generate shorter hash codes,
since more data-specific information can be exploited.
Since the space spanned by meaningful images is in general only a small portion of the entire vector space,
by using a machine-learning strategy,
it is possible to tailor a hashing scheme to cater to this space so that more compact binary codes may be obtained (\ie the images can be represented by shorter binary hash codes). 
Several hashing techniques in this category, like~\cite{cnnh,ksh,mlh,dgh}, have been proposed,
reporting promising performance.
These techniques can be further divided into unsupervised approaches and supervised approaches.
Unsupervised approaches use unlabeled training data to learn the hash functions.
Representative algorithms include PCA hasing~\cite{pcah},
which is based on principal component analysis (PCA);
Iterative Quantization (ITQ)~\cite{itq},
which applies orthogonal rotation matrices to tune the initial projection matrix learned by PCA;
and Spectral Hashing (SH)~\cite{sh},
which is based on the eigenvectors computed from the data similarity graph.
Deep Hashing (DH)~\cite{sdh} is another example, which leverages the capacity of neural networks on acquiring better visual features for improved performance.

With training data that come with label information,
such as point-wise labels,
pairwise labels~\cite{cnnh, dsh},
or triplet labels~\cite{dnnh},
supervised methods have been developed to take advantage of the extra information.
Well-known supervised approaches include Supervised Hashing with Kernels (KSH)~\cite{ksh},
which learns the hash function in a kernel space;
Minimal Loss Hashing (MLH)~\cite{mlh},
which minimizes a hinge loss function to learn the hash function;
Binary Reconstructive Embeddings (BRE)~\cite{bre},
which learns hash functions by minimizing the reconstruction error between the vectors from the original space and the Hamming space;
and Supervised Deep Hashing (SDH)~\cite{sdh}, which learns the binary codes by a deep neural network.
While these methods use relaxation schemes to obtain the discrete binary codes,
Discrete Graph Hashing (DGH)~\cite{dgh} and Supervised Discrete Hashing (SDiscH)~\cite{sdish} were also proposed to calculate the optimal binary codes directly,
and improved performance was reported.


The afore-mentioned approaches in general treat the feature extraction step and the binary encoding step as two different stages. Recently, some deep-learning-based approaches~\cite{cnnh, dnnh, dsh, dpsh, sdh, dph} have been proposed, which attempt to learn the binary representation simultaneously with the features by using convolutional neural networks (CNNs).
When the learned hash codes are used for tasks like image retrieval,
these recent methods have been shown to improve the performance quite significantly.
Reviewing these approaches,
one may realize that the feature-learning stage is global in nature,
although local information (like saliency or region interaction) may be essential for tasks like retrieving images containing foreground objects of similar type (but possibly with diverse background).

In existing studies, performance comparison among different hashing models has been based on metrics that do not necessarily reflect the advantages of the hashing-based scheme. For example, in Fig. \ref{fig:usage},
we plot the histogram of the hash codes of some existing hashing models, showing that importing label information in the objective function may lead to higher mAP scores, but does not necessary increase the utilization of the hash space, and thus may move the retrieval problem towards a classification-like problem.
This motivated us to develop a new performance metric that considers not only the retrieval accuracy
but also the utilization the hash space.



\section{Performance metrics for Retrieval} \label{sec:metric}
Let $\mathbb{U}$ be the sample space. A good hashing scheme aims to learn a mapping $f: \mathbb{U} \rightarrow \{-1,1\}^k$,
such that for any $u_1, u_2 \in \mathbb{U}$, if $u_1, u_2$ are similar,
then $b_1, b_2$ are similar,
where $b_1 = f(u_1)$ and $b_2 = f(u_2)$.
After learning such mapping,
for a given input query $u$,
we obtain their binary representation $b = f(u)$,
and compare $b$ with the other binary code of the database by using the Hamming distance $d_H(\cdot, \cdot)$.

We now describe several commonly-used metrics for image retrieval, and discuss the reason why they are not suitable for hashing-based models. Note that we only discuss score-based metrics, and plot-based metrics like Precision-Recall curves are not considered here.




\subsection{Precision}
For large-scale datasets on the Internet, recall becomes a less important score of an image retrieval system,
as few users will be interested in searching {\sl all} images relevant to a query.
On the other hand, precision is of importance, 
which is defined as:
\begin{equation}
    P = \frac{\text{number of true positives}}{\text{number of true positives + number of false positives}}
\end{equation}
Usually, retrieval precision is measured in two ways:
(1) based on the top $K$ returned samples ($\ie$ $P@K$);
(2) based on the samples having a Hamming distance $\leq r$ to the query ($\ie$ $P@d_H \leq r$).
Although precision can reflect the performance of a hashing model in some degree,
it does not consider the ranking order of the returned samples.
For instance, for a given query, we have the same value $P@2$ for the following two retrieval results:
\begin{equation}\label{case1}
    \begin{split}
        &\text{The top $1st$ and $2nd$ returns are true positive}\\ 
        &\text{and false positive respectively}
    \end{split}
\end{equation}
\begin{equation}\label{case2}
    \begin{split}
        &\text{The top $1st$ and $2nd$ returns are false positive}\\
        &\text{and true positive respectively}
    \end{split}
\end{equation}
Hence the goal of keeping relevant images on the top as much as possible cannot be guaranteed by simply pushing for better $P@K$ values.

\subsection{Mean Average Precision}
Due to the disadvantage of using precision, the Mean Average Precision (mAP) has become a much more popular metric for measuring the performance of retrieval systems.
The mAP of a sequence of queries is the mean of the average precision of each query.
To be precise, we first give the definition of Average Precision (AP) of the top $K$ returns:
\begin{equation}
    AP(b_j)@K
    = \frac
    {\sum_{i=1}^k \mathbb{I}(b_j) P(b_j)@i}
    {\sum_{i=1}^k \mathbb{I}(b_j)}
\end{equation}
where $\mathbb{I}$ is the indicator function:
\begin{equation}
    \mathbb{I}(b) = 
    \begin{cases}
    1   &\text{if b is a true positive}\\
    0   &\text{if b is a false positive}
    \end{cases}
\end{equation}

The mAP can be calculated by taking the mean of AP of a query set $B = \{b_1, ..., b_N\}$:
\begin{equation}
    mAP(B) = \frac{\sum_{j=1}^N AP(b_j)@K}{N}
\end{equation}
Therefore, the mAP scores for the two situations we presented in the previous subsection could be mAP$=1$ for case \ref{case1} and mAP$=0.5$ for case \ref{case2} respectively. mAP could be better than Precision $@k$ in demonstrating the ranking order of retrieved images.
For the above metrics,
we call them \textbf{global} if all the data in the database take parts during calculation,
otherwise we call them \textbf{local}.
So $P@k$ and $P@d_H \leq r$ are usually used as local metrics,
while mAP can be either local or global, depending on whether it is calculated on the top $K$ returns or the whole dataset.

\subsubsection{Uniqueness of mAP}
\begin{figure}
	\begin{center}
		\includegraphics[width=1\linewidth]{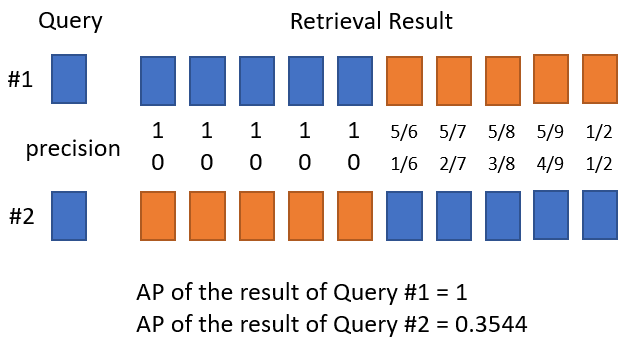}
	\end{center}
	\caption{
	    The figure shows the different retrieval results of the same query.
	    There are two classes, the samples in class 1 are denoted by blue,
	    while the samples in class 2 are denoted by orange.
	    Suppose there are 10 samples in the database,
	    all of them are mapped to the same binary code by some learned hashing model,
	    then the retrieval result is not unique.
	    As shown in the figure,
	    the average precision of the best and the worst result are 1 and 0.3544 respectively.
	}
	\label{fig:unique}
\end{figure}
Even though the mAP can reflect the ranking order of retrieved images, for a hashing-based retrieval model, the mAP score of a given testing set is not unique in general.
When there are collisions in one binary code ($\ie$ two or more samples from different classes are mapped to the same binary code), there can be several ways to arrange the retrieval ranking order since the Hamming distances of the retrieved images are the same to the query images.
The binary code collision will lead to different mAP scores for the same query image.
Fig. \ref{fig:unique} shows the best case and worst case for a two-class example.
This phenomenon suggests that mAP could be very misleading when it is used to compare two models that have very high collision rate in the hash space.

\subsection{Usage of Hash code}
For hashing-based approaches, to take advantage of the binary space, we would want to utilize the hash space well so as to reduce collisions, which would in turn facilitate retrieval ($e.g.$, supporting more refined ranking). 
However, the commonly used metrics mentioned above do not consider this issue of utilization of the hash space.
Even worse, since high mAP means that the samples from the same class are ranked high, meaning that the distances among those samples are small or even zero, it may lead to high collision. In the extreme case, this could push the problem towards a simple classification problem, as we have discussed previously.
In the following, we analyze this issue further and show why we should avoid using a global metric.
We show that achieving a high score for such a metric may lead to low utilization of the hash space.
\begin{figure}
	\begin{center}
		\includegraphics[width=0.8\linewidth]{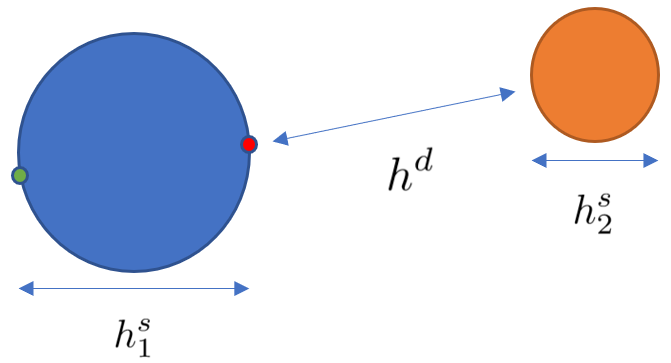}
	\end{center}
	\caption{
	    The figure shows the relation between $h_j^s$ and $h_j^d$.
	    Assume $j = 1$(blue) or $2$(orange),
	    to achieve mAP = 1,
	    we have $h^d > \max(h_1^s, h_2^s)$,
	    as for the input query(red point),
	    the farthest binary code in the same class (green point) should be ranked higher than any point in class $2$.
	}
	\label{fig:dist}
\end{figure}
\begin{figure}
	\begin{center}
		\includegraphics[width=0.8\linewidth]{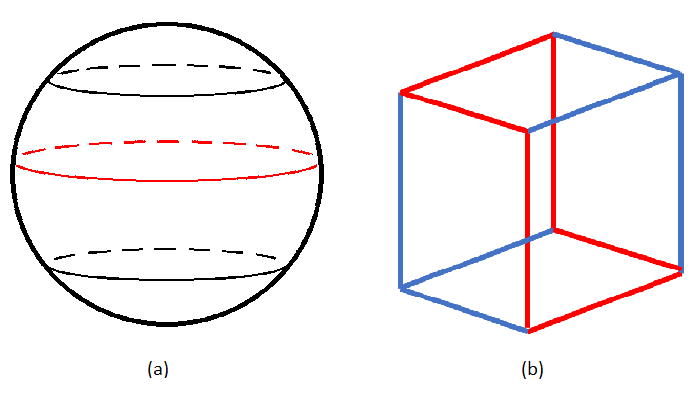}
	\end{center}
	\caption{
	    The figures show the concept of orthodrome,
	    which is also known as great circle.
	    (a) The orthodrome on a sphere,
	    which is the intersection of the sphere and a plane which passes through the center of the sphere,
	    is indicated by the red line.
	    (b) The orthodrome on a cube, which is indicated by the red line.
	}
	\label{fig:ortho}
\end{figure}

\begin{prop}
For a hashing-based retrieval system, a perfect global mAP value (or other global metric) implies that the utilization the hash space is less than or equal to $2/3$.
\end{prop}
Assume that there are $N$ classes in the dataset.
To achieve mAP $=1$,
for any input query $b \in \{-1,1\}^k$,
the retrieval system must rank the samples in the same class the highest.
So, it is natural to assume that,
the samples in the $N$ classes,
through the hash functions,
are mapped to $N$ sets $\{\Delta_j\}_{j=1}^N$ in a $k$-bit binary space separately.
For each $\Delta_j$, 
we define two distances:
\begin{align}
    h^s_j &= \max \{d_H(a,b) \mid \forall a,b \in \Delta_j\}\\
    h^d_j &= \min \{d_H(a,b) \mid \forall a \in \Delta_j, \forall b \in \Delta_i, i\neq j \}
\end{align}
where $h^s_j$ denotes the diameter of $\Delta_j$,
and $h^d_j$ denotes the shortest distance from $\Delta_j$ to another Hamming ball.
By requiring $h^s_j < h^d_j$, we can achieve mAP $= 1$.
To prevent every samples in a class from being mapped to a single binary code,
we further assume $\tilde{h^s} = min_j h^s_j > 0$.
The illustration is demonstrated in Fig. \ref{fig:dist}.

\begin{figure}
	\begin{center}
		\includegraphics[width=0.9\linewidth]{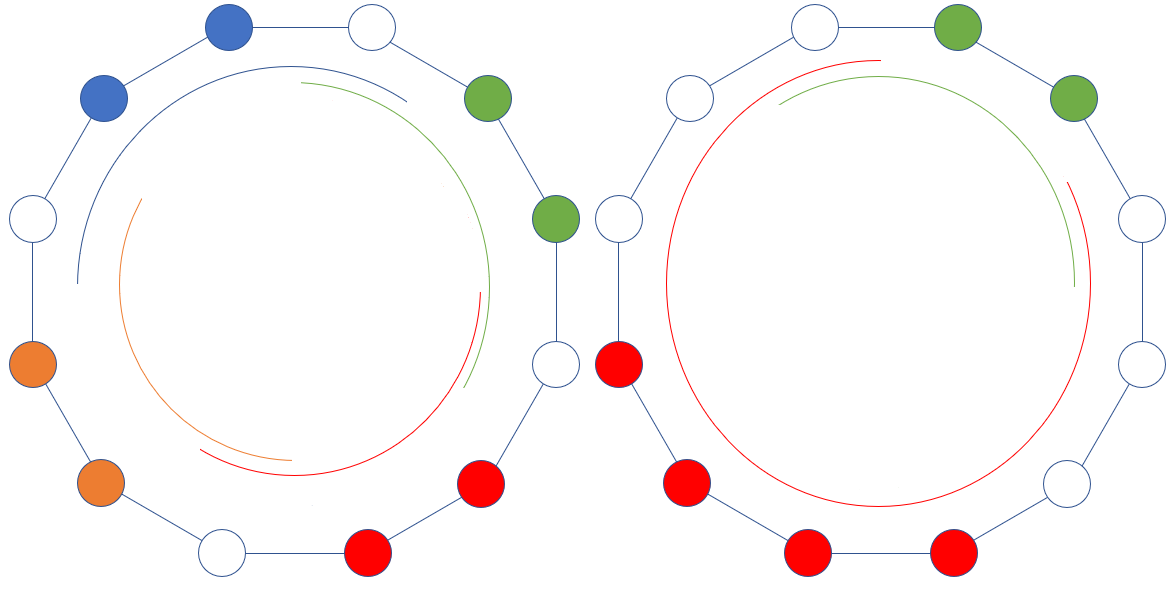}
	\end{center}
	\caption{
	    Two orthodrome on a 12-bit binary space.
        Binary codes belonging to the same class are represented by the same color.
        The blue arc inside the orthodrome means that the corresponding nodes cannot come from a class other than the blue class.
        To maintain mAP = 1,
        the same restriction applies to the green, red and orange arcs.
        When $\tilde{h^s} = 1$:
        (left) to attain the maximal usage,
        all the $h^s_j$ should be 1;
        (right) an example that if $h_{red}^s$ is at least 3.
	}
	\label{fig:circle}
\end{figure}

To prove that the utilization of the hash space will be less than $2/3$,
we have to define the orthodrome on the binary space $\{-1,1\}^k$.
For any $b \in \{-1,1\}^k$,
we can flip the elements from either $-1$ to $1$ or $1$ to $-1$.
If we flip all the elements of $b$ one at a time,
we obtain a path from $b$ to its farthest point $\overline{b}$ in this space.
Following the same flipping order,
we obtain another path from $\overline{b}$ to $b$.
An orthodrome contains all the points on this two paths.
For example,
if $b = [-1,-1,-1,-1]$,
then its farthest point $\overline{b}$ will be $[1,1,1,1]$.
If we flip the sign of the elements of $b$ in the order of $2,4,1,3$,
then the path from $b$ to $\overline{b}$ and back to $b$ can be stated as follows:
\begin{equation}
\begin{split}
    &[-1,-1,-1,-1]
    \rightarrow
    [-1,1,-1,-1]\\
    \rightarrow
    &[-1,1,-1,1]
    \rightarrow
    [1,1,-1,1]\\
    \rightarrow
    &[1,1,1,1]
    \rightarrow
    [1,-1,1,1]\\
    \rightarrow
    &[1,-1,1,-1]
    \rightarrow
    [-1,-1,1,-1]\\
    \rightarrow
    &[-1,-1,-1,-1]
\end{split}
\end{equation}
Then these points form an orthodrome.
An example of orthodrome is provided in Fig. \ref{fig:ortho}.
If on an orthodrome, there is a binary code belonging to class $j$,
since $h^s_j < h^d_j$,
its closest $h^s_j$ codes cannot come from the class other than class $j$
As this statement is true for any class $j$,
the utilization of the binary codes on this orthodrome must be less than or equal to $2/3$.
(See Fig. \ref{fig:circle} for illustration).
As this holds for any orthodrome,
we can conclude that the utilization of the hash codes in the entire space is not greater than $2/3$.
Note that the upper is not the supremum ($i.e.$ the least upper bound),
and if we decrease the number of classes $N$ and increase $\tilde{h^s}$,
which is the lower bound of the diameter of the $\delta_j$,
the upper bound for the utilization of the hash space will drop.

\section{The Proposed Metric}\label{sec:proposed}
\begin{figure}
	\begin{center}
		\includegraphics[width=0.9\linewidth]{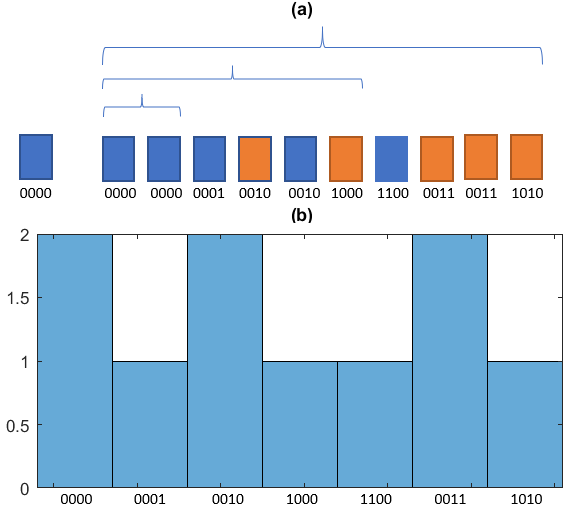}
	\end{center}
	\caption{
        The figures show the illustration on how to compute LGAP.
        (a) An input query $b$ and result of $P@d_H \leq 2$.
        The three groups represent $S_0(b)$, $S_1(b)$ and $S_2(b)$ respectively.
        (b) An example of how to calculate the penalty function.
        Here, $\varphi(S_1(b)) = $
        Area of all rectangles / (Area of the largest Rectangle $\times$ number of bins).
	}
	\label{fig:lgap}
\end{figure}

In this section, we define our proposed metric for hashing-based retrieval system.
For a given set of query $B = \{b_1, ..., b_N\}$,
$b_j \in \{-1,1\}^k$ $\forall j \in \{1, ..., N\}$, and a set of binary codes of relevant images $X = \{x_1, ..., x_M\}$, $x_i \in \{-1,1\}^k$ $\forall i \in \{1, ..., M\}$. We can define the set of retrieved binary codes of certain query $b_j$ as $S_r(b_j) = \{x_i \mid d_H(b_j,x_i) \leq r, \forall i \in \{1, ... M\} \}$, where the $d_H(\cdot,\cdot)$ denotes the Hamming distance.
Then the Local Group Average Precision can be calculated as follows:
\begin{equation}
    LGAP(b_j)@r = \frac{\sum_{k=0}^r (P@d_H \leq k) (\varphi(S_k(b_j))} {r+1}
\end{equation}
where $\varphi$ is a penalty function,
which maps the the usage of binary codes of subset $S_r(b_j)$ to $[0,1]$.
The purpose of setting the penalty term is to encourage the dispersion of the binary codes.
If for a query sample,
the usage of the binary codes inside a Hamming ball is not disperse enough,
the output of $\varphi$ should be very small and get close to $0$,
if the binary codes spread out in the Hamming ball (uniformly distributed at best),
the output of $\varphi$ should be large and get close to $1$.
$\varphi$ can be any function fulfill these requirements.
In our experiments, we define $\varphi$ as
    $\varphi(S) = \frac{A(S)}{B(S)}$
where $A(S)$ is the summation of histogram of each binary code within a Hamming ball and $B(S)$ is the product of the largest histogram and the number of binary codes in the Hamming ball ($\ie$ the rectangular area covering all histograms). The Fig. \ref{fig:lgap} part (b) is the illustration of how to compute our penalty function. The $A(S)$ equals to the blue shade area in the part (b) and $B(S)$ is the total rectangular area which contains all histogram. Therefore, if the histogram is uniformly distributed in the Hamming ball ($i.e.$ each binary code encodes same number of images), the $\frac{A(S)}{B(S)}$ should equal to $1$. 


Similar to mAP,
the Mean Local Group Average Precision can be calculated by taking the mean of the LGAP of the query set:
\begin{equation}
    mLGAP(B)= \frac{\sum_{j=1}^N LGAP(b_j)}{N}
\end{equation}
Therefore, the $LGAP(b)@2$ for the example in Fig. \ref{fig:lgap} can be calculated as:
\begin{align}
    &LGAP(b)@2 \\
    = &\frac{\sum_{k=0}^2 (P@d_H \leq k) (\varphi(S_k(b_j)))} {3}\\
    = &\frac{(1)(1) + (\frac{4}{6}) (\frac{6}{2(\binom{4}{0}+\binom{4}{1})})
    + (\frac{5}{10}) (\frac{10}{2(\binom{4}{0}+\binom{4}{1}+\binom{4}{2})})} {3}\\
    = &0.5424
\end{align}

\subsection{Advantages of uniformly distributed hash codes}
Traditional hash functions are used in hash tables, where a good hash function should map the inputs as uniformly as possible to the hash table so as to minimize collision, which in turn minimizes search time. While this property remains certainly desired in hashing-based retrieval, there is another advantage for making the hash codes distributed uniformly. When hashing techniques are used for retrieval, they are often used to approximate nearest neighborhood search. We would expect that, if we have a good hash model, then for similar hash codes, the corresponding samples may have similar properties in the original domain (e.g., visual or semantic domains).
For images, even if they are in the same class, some of them may be more similar than other others. Hence using the hash space as uniformly as possible can leave more room for allowing subtle distinctions of images in the hash space.
\begin{figure*}
	\begin{center}
		\includegraphics[width=0.7\linewidth]{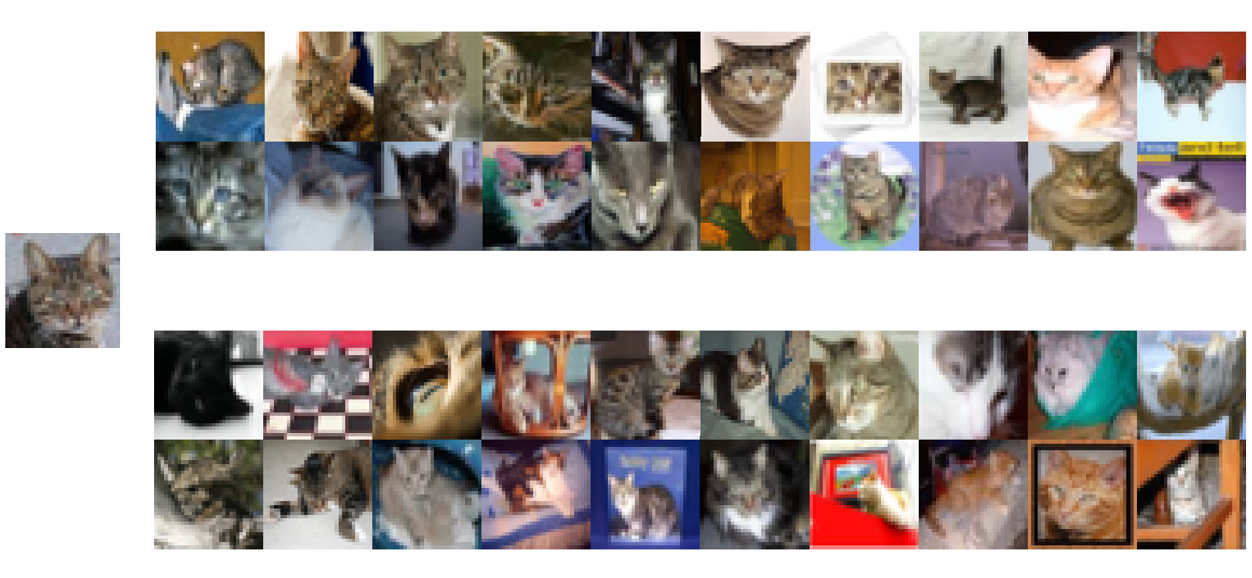}
	\end{center}
	\caption{
	    The retrieval result for a given query "cat head".
	    Only true positives are included.
	    The first and second rows are the results for $d_H \leq 1$;
	    The third and forth rows are the results for $d_H = 2$.
	    The result shows that "cat head" appear in $ d_H \leq 1$ more frequently.
	}
	\label{fig:ex}
\end{figure*}

In Fig. \ref{fig:ex} we show an visual example from our proposed objective function, which is discussed in Section \ref{sec:exp}. The retrieval system tends to map samples to different hash codes while preserving some similarity:
for the given query ``cat head", the retrieved ``cat head" are mainly distributed in Hamming radius $\leq 1$.
This is a clear advantage over existing popular metrics that do not consider collision.

\section{Experiments}\label{sec:exp}

\begin{table*}[]
\centering
\caption{
    The mLGAP and mAP score for DSH under different objective functions:
    (1) Eq.\ref{eq:dsh} and \ref{eq:new10} for CIFAR-10.
    (2) Eq.\ref{eq:dsh} and \ref{eq:new100} CIFAR-100.
}
\label{tab:cifar}
\begin{tabular}{|c|c|c|c|c|c|c|c|c|}
\hline
    & \multicolumn{4}{c|}{CIFAR10}                              & \multicolumn{4}{c|}{CIFAR100}                             \\ \hline
    & \multicolumn{2}{c|}{12-bit} & \multicolumn{2}{c|}{24-bit} & \multicolumn{2}{c|}{12-bit} & \multicolumn{2}{c|}{24-bit} \\ \hline
                                    & mLGAP         & mAP       & mLGAP         & mAP         & mLGAP         & mAP         & mLGAP         & mAP         \\ \hline
\ref{eq:dsh}                        & 0.2633        & 0.6012    & 0.2889        & 0.6407      & 0.0637        & 0.1514      & 0.0660        & 0.1556      \\ \hline
\ref{eq:new10} \& \ref{eq:new100}   & 0.2796        & 0.3008    & 0.3461        & 0.3124      & 0.1703        & 0.1156      & 0.3310        & 0.1187      \\ \hline
\end{tabular}
\end{table*}
We evaluate our proposed mLGAP metric on CIFAR-10 and CIFAR-100\cite{cifar10} datasets. 
The results show that our proposed metric has the capability of reflecting both retrieval accuracy and dispersion of hash codes, hence potentially providing a better performance measure for evaluating hashing-based retrieval. 

\subsection{Dataset}
Both CIFAR10 and CIFAR-100 datasets are used to verify the effectiveness of our proposed dispersion scheme in objective function, which will be described in details later,  for compact hashing function learning and compare our results with those from several state-of-the-art online hashing approaches with the proposed new mLGAP metric.

\noindent \textbf{CIFAR-10:}
The CIFAR-10 dataset~\cite{cifar10} contains 10 mutually exclusive categories with 6,000 color images in each category, in total 60,000 color images of size 32 $\times$ 32.
Officially there are 5000 training images and 1000 testing images per class and we follow the training and testing splits to train our compact hashing codes learning approaches.

\noindent \textbf{CIFAR-100:}
The CIFAR-100 dataset~\cite{cifar10} is similar to CIFAR-10 dataset, which totally contains 60,000 color images with size 32 $\times$ 32, but only 600 color images for each category since there are 100 categories in total. Additionally, each 5 relevant categories assemble to form a new superclass, thus 20 superclasses in total. In our experiment, we consider both "fine" labels (the class which each image belongs to) and "coarse" labels (the superclass which each image belongs to) in training stage and in testing stage, performance measurement is only based on "coarse" labels.

\subsection{Dispersion Scheme}
In our experiments, we utilize the DSH model from ~\cite{dsh}, which achieves the state-of-the-art performance on CIFAR-10 dataset, to learn the hash codes. However, the DSH model is a supervised learning approach, whose disadvantages for retrieval has been discussed in Section \ref{sec:related}. Therefore, while keeping the same network structure as the DSH model, we propose a new objective function so that the learned hash codes can be dispersed as much as possible. 

The original objective function of DSH model can be presented as following:
\begin{equation}\label{eq:dsh}
\mathcal{L}_{\mathcal{r}}  	
= \sum \{ \frac{1}{2} (1 - y_i) A
+ \frac{1}{2} y_i B 
+ \alpha C \}
\end{equation}
where
\begin{align}
    A &= \|b_{i,1} - b_{i,2}\|_2^2 \\
    B &= \max(m - \|b_{i,1} - b_{i,2}\|_2^2, 0) \\
    C &= \|b_{i,1} - 1\|_1 + \|b_{i,1} - 1\|_1
\end{align}
$y_i = 0$ if two images are from the same class and $y_i = 1$ otherwise; $m$ is the margin value for calculating the dissimilarity of two binary-like codes. The first two terms are designed to make the hash codes more similar if they come from the same category and more different otherwise. The last term is the regularizer which is utilized as forming the binary-like continuous output codes close to $\{-1,1\}$.

We can observe that, except the regularizer, the other two terms use L2 Norm to compute the distance between binary-like codes. However, calculating the Hamming distance is similar to the calculation the L0 Norm (that is, for any vector $a$ and $b$ having the same length, the L0 Norm equals to the number of different elements of $a$ and $b$). Since L1 relaxation approximates the L0 Norm better than L2 Norm, we utilize L1 instead of L2, which leads to an improvement over the original objective function.

The new objective function is also designed based on the distance in the Hamming space. In the original objective function, if two images are from the same category, then two binary-like output codes should be pushed as much close as possible ($i.e.$ the first term in eq. \ref{eq:dsh}). This will certainly exacerbate the collision problem. Therefore, we design a buffer zone for learning the hash codes for the same category. We do not necessarily push the binary-like outputs from the same category as close as possible, but within a certain range the two binary-like outputs can be defined as the same category. 

So the new objective function for training single label datasets ($i.e.$ CIFAR-10 dataset) can be redesigned as following, factoring the aforementioned improvements:
\begin{equation}\label{eq:new10}
\mathcal{L}_{\mathcal{r}}  	
= \sum \{ \frac{1}{2} (1 - y_i) A
+ \frac{1}{2} y_i B
+ \alpha C \}
\end{equation}
where
\begin{align}
    A &= \max(r_1 - \|b_{i,1} - b_{i,2}\|_1, 0)\\
    &+ \max(\|b_{i,1} - b_{i,2}\|_1 - r_2, 0)\\
    B &= \max(m - \|b_{i,1} - b_{i,2}\|_1, 0)\\
    C &= \|b_{i,1} - 1\|_1 + \|b_{i,1} - 1\|_1
\end{align}
$0 < r_1 < r_2$ is the distance range for the binary-like codes from the same category. 

For CIFAR-100, which contains both classes and superclasses, we design two buffer zones for learning the binary codes. The distant range for codes from the same superclass is larger than the distant range for codes from the same class, implying that the variance of the binary-like codes from the same class should be less than that of the same superclass. Therefore, the new objective function for CIFAR-100 can be written as:
\begin{equation}\label{eq:new100}
\mathcal{L}_{\mathcal{r}}  	
= \sum \{ \frac{1}{2} (1 - y_i)A 
+ \frac{1}{2} (1 - Y_i)B 
+ \frac{1}{2} Y_i C
+ \alpha D \}
\end{equation}
where
\begin{align}
    A &=\max(r_1 - \|b_{i,1} - b_{i,2}\|_1, 0) \\ 
    &+ \max(\|b_{i,1} - b_{i,2}\|_1 - r_2, 0) \\
    B & =\max(r_3 - \|b_{i,1} - b_{i,2}\|_1, 0) \\ 
    &+ \max(\|b_{i,1} - b_{i,2}\|_1 - r_4, 0) \\
    C &= \max(m - \|b_{i,1} - b_{i,2}\|_1, 0)\\
    D &= \|b_{i,1} - 1\|_1 + \|b_{i,1} - 1\|_1
\end{align}
$Y_i = 0$ if two images are from the same superclass, and $Y_i = 1$ otherwise. $0 < r_1 < r_2 \leq r_3 < r_4$ is the two distance ranges for the same class and same superclass respectively. 

We applied the new objective function with the model proposed in ~\cite{dsh} which contains 3 convolutional layers and 2 fully connected layers. The performance on the new mLGAP metric will be presented in the following section.

\subsection{Evaluation and Parameter Settings}
\begin{figure}
	\begin{center}
		\includegraphics[width=1\linewidth]{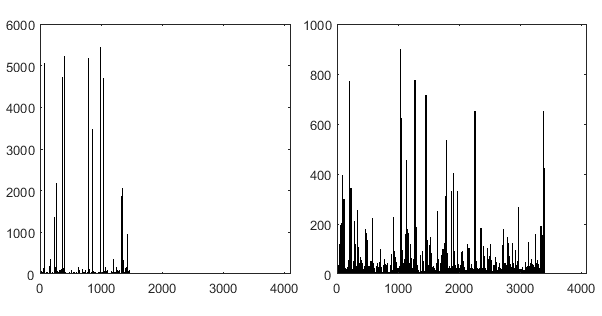}
	\end{center}
	\caption{
        The figure shows the usage of the 12-bit hash codes on CIFAR-10 from different models.
        (Left) DSH with the original objective function;
        (Right) DSH with our proposed objective function.
	}
	\label{fig:usage2}
\end{figure}
In our experiments, all the margin values like $m$, $r_1$, $r_2$, $r_3$ and $r_4$ are all set heuristically. In the $k$-bit binary space, we set $m = 2k$ to encourage dissimilar images at least $\frac{k}{2}$~\cite{dsh} apart. Set $r_1 = \frac{k}{6}$ and $r_2 = \frac{k}{3}$ for the objective function in training CIFAR-10 dataset. The distance range $r_1 = \frac{k}{12}$, $r_2 = \frac{k}{6}$, $r_3 = \frac{k}{6}$ and $r_4 = \frac{k}{3}$ are set for training with CIFAR-100 dataset. All training stage for both CIFAR-10 and CIFAR-100 dataset contains 750 epochs and learning rate decays $40\%$ every 150 epochs from 0.01 at the beginning. The model is trained and updated by batches with batch size$=200$.
Fig. \ref{fig:usage2} shows the usage of 12-bit binary codes of the hashing model on CIFAR-10.
The final performance is listed in Tab. \ref{tab:cifar}, suggesting that the new metric reflects not only the performance but also the uniformity of usage of the binary codes.


\section{Conclusions}\label{sec:conc}
In this paper we discussed the commonly-used metrics for hashing-based retrieval systems.
We present a new metric which not only consider the retrieval accuracy but also the utilization of the hash space,
hence providing a better performance measure of evaluating hashing-based retrieval.
We show, by experiments, when the samples are mapped more uniformly to the binary space,
the hash code can preserve some similarity of the original samples, in the sense of either visual or semantic meaning.

{\small
\bibliographystyle{ieee}
\bibliography{egbib}
}

\end{document}